\documentclass{article}

\usepackage{arxiv}

\usepackage[utf8]{inputenc} 
\usepackage[T1]{fontenc}    
\usepackage{hyperref}       
\usepackage{url}            
\usepackage{booktabs}       
\usepackage{amsfonts}       
\usepackage{nicefrac}       
\usepackage{microtype}      
\usepackage{lipsum}
\usepackage{graphicx}
\usepackage{multirow}
\graphicspath{ {./images/} }

\usepackage{authblk}
\usepackage{listings}
\usepackage{enumitem}

\begin{document}

\title{Pingan-vcgroup’s solution for icdar
2021 competition on scientific table image recognition to latex}

\author[1]{Yelin He}
\author[1]{Xianbiao Qi}
\author[1]{Jiaquan Ye}
\author[2]{Peng Gao}
\author[1]{Yihao Chen}
\author[1]{Bingcong Li}
\author[1]{Xin Tang}
\author[1]{Rong Xiao}
\affil[1]{Visual Computing Group, Ping An Property \& Casualty Insurance Company}
\affil[2]{Ping An Technology Company}

\thanks{Xianbiao Qi is the corresponding author. If you have any questions or concerns about the implementation details, please do not hesitate to contact heyelin58@gmail.com or qixianbiao@gmail.com.}

\maketitle
\begin{abstract}
This paper presents our solution for the ICDAR 2021 Competition on Scientific Table Image Recognition to LaTeX. This competition has two sub-tasks: Table Structure Reconstruction (TSR) and Table Content Reconstruction (TCR). We treat both sub-tasks as two individual image-to-sequence recognition problems.
We leverage our previously proposed algorithm MASTER~\cite{lu2019master}, which is originally proposed for scene text recognition. We optimize the MASTER model from several perspectives: network structure, optimizer, normalization method, pre-trained model, resolution of input image, data augmentation, and model ensemble. 
Our method achieves 0.7444 Exact Match and 0.8765 Exact Match @95\% on the TSR task, and obtains 0.5586 Exact Match and 0.7386 Exact Match 95\% on the TCR task.
\end{abstract}



\section{Introduction}
Recognizing a table image into a Latex code is challenging due to complexity and diversity of table structures and long sequence problem compared to traditional OCR. The challenge aims at assessing the ability of state-of-the-art methods to recognize scientific tables into LaTeX codes. In this competition, there are two sub-tasks with different levels of difficulty.\\\\
\textbf{Subtask I Table Structure Reconstruction} is to reconstruct the structure of a table image into the form of LaTeX code but ignore the content of the table.\\\\
\textbf{Subtask II Table Content Reconstruction} is to reconstruct the structure and the content of a table image simultaneously into the form of LaTeX code.\\\\
Table Image Recognition to LaTeX data set~\cite{Pratik2021ICDAR} is a scientific table recognition dataset which consists of 43,138 training samples, 800 validation samples, 2,203 test samples on the TSR task, and 35,500 training samples, 500 validation samples, 1,917 test samples on the TCR task. We treat both two tasks as an image-to-sequence recognition task as scene text recognition. Our model is based on our previous work MASTER~\cite{lu2019master}, and it performs well on OCR task and can be freely adopt to other similar tasks, such as curved text prediction, multi-line text prediction, vertical text prediction, multilingual text prediction. The rest of the paper is organized as follows. We analyze the competition data set in Section~\ref{sec:datasec}, introduce the method and the tricks on the TSR and the TCR tasks in Section~\ref{sec:methodsection}, and then present our experimental results in Section~\ref{sec:experiments}, and finally conclude the paper in Section~\ref{sec:conclusion}.

\section{Data}
\label{sec:datasec}
In this section, we conduct some statistical analysis of the provided data. Our choices of some model parameters are based on the analysis obtained in this section. 

\subsection{Statistics of Data}
\begin{figure}[h] 
    \centering
    \includegraphics[width=0.75\textwidth]{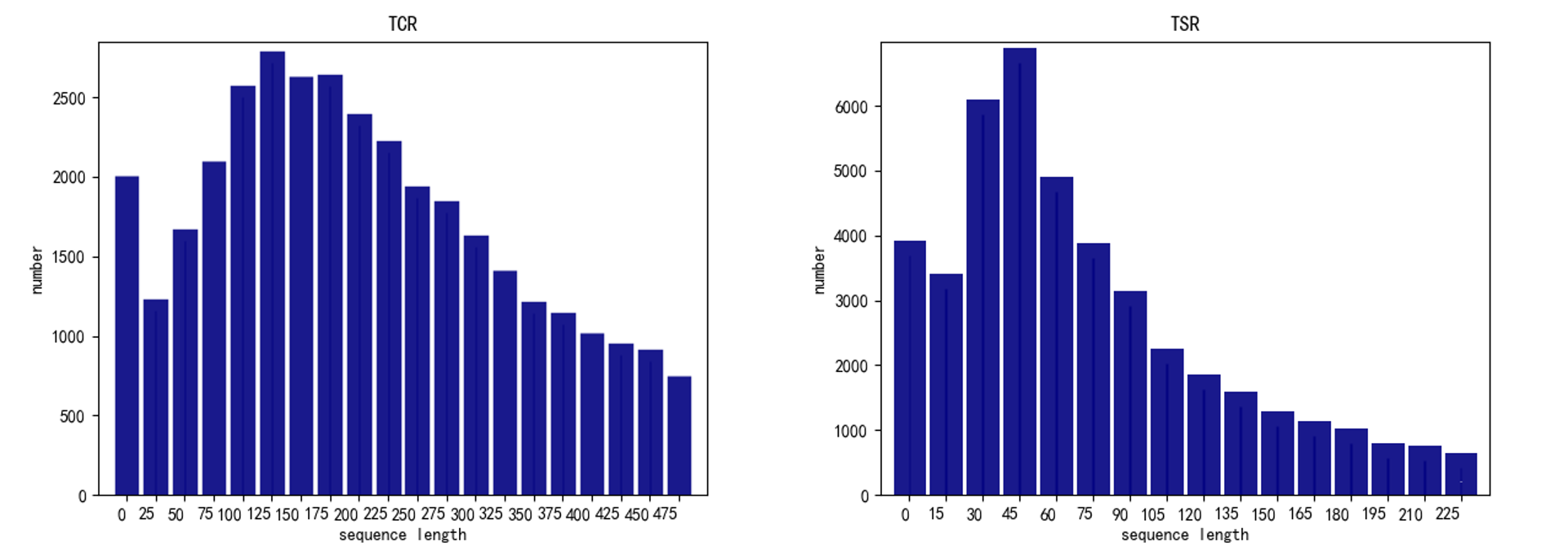}
    \caption{Distribution of sequence length. The maximum sequence lengths are 500 for the TCR task, and 250 for the TSR task.}
    \label{fig:sequencelength}
\end{figure}

The sequence length distribution of TSR data set and TCR data set are shown at Figure~\ref{fig:sequencelength}. As shown in Figure~\ref{fig:sequencelength}, Tables have less than 250 tokens for the TSR task and less than 500 tokens for the TCR task. The distribution of sequence lengths appears relatively uniform; the average numbers of sequence lengths are 76.09 for the TSR task and 213.95 for the TCR task separately. 
Thus, in our MASTER model, the maximum sequence length is set to be 500 for the TCR task, and  250 for the TSR task. 

\begin{figure}[h] 
    \centering
    \includegraphics[width=0.75\textwidth]{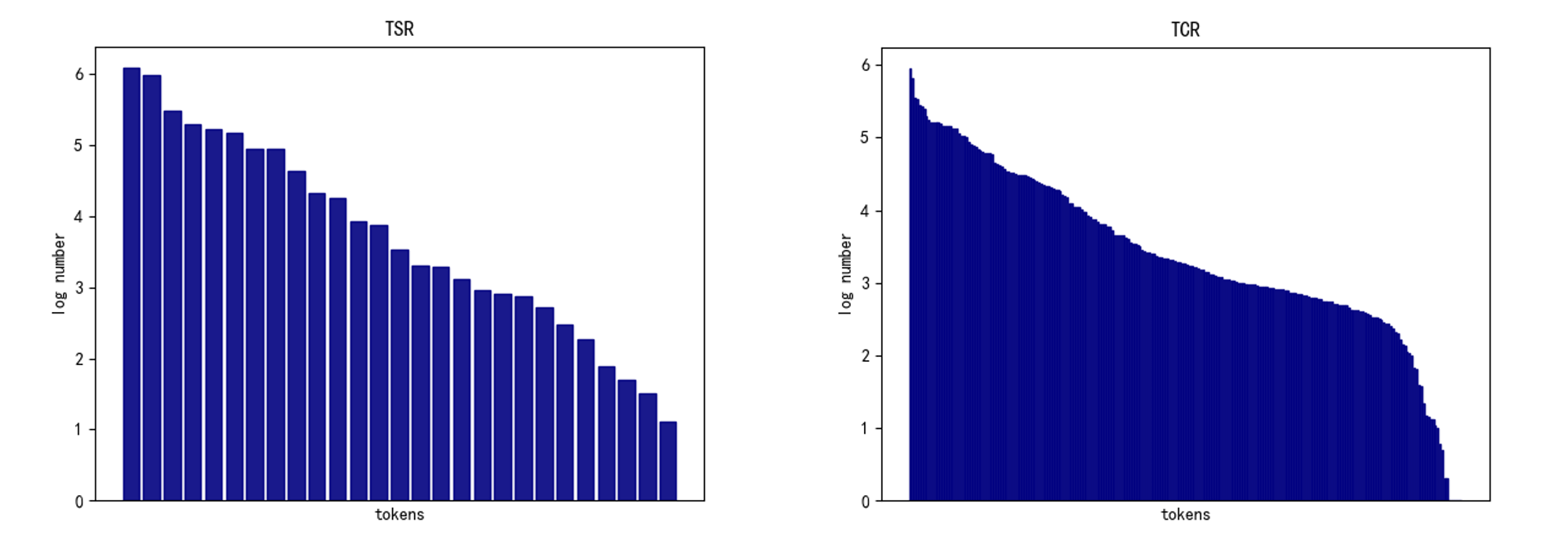}
    \caption{Distribution of tokens number. For aesthetic display, we apply the logarithmic function to the distribution. The distribution of tokens varies largely.}
    \label{fig:tokennumber}
\end{figure}

The tokens number distribution of TSR data and TCR data are shown at Figure~\ref{fig:tokennumber}. There are only 27 tokens in the TSR task; it shows extreme imbalance in distribution of the tokens. The most frequent token is ``CELL'' that appears 1,217,487 times, while the least token appear only several dozen of times. Similarly, the category imbalance also appears in the TCR task. Specifically, all special mathematical characters are represented by ``LATEX\_TOKEN'', so there are only 236 tokens in the TCR task. The tokens ``2c'', ``2ex'', ``564'' and ``constant'' only appear one time so that we believe that there are some mislabeling; we therefore replace them with ``LATEX\_TOKEN''.

\section{Method}
\label{sec:methodsection}
Overall, our model is based on our previous work MASTER~\cite{lu2019master}. We recommend you to refer to the original manuscript for detailed information. In this section, we will mainly describe some useful attempts we conduct in this competition.



\subsection{Task1: Table Structure Reconstruction}
In the TSR task, we mainly use the following strategies to improve accuracy.\\\\
\textbf{Ranger Optimizer.} Ranger~\cite{lessw2019ranger} integrates RAdam (Rectified Adam)~\cite{liu2019radam}, LookAhead~\cite{zhang2019lookahead}, and GC (gradient centralization)~\cite{yong2020gradient} into one optimizer. LookAhead can be considered as an extension of Stochastic Weight Averaging (SWA)~\cite{izmailov2018averaging} in the training stage.\\\\
\textbf{Data Augmentation.} Our data augmentation method is based on~\cite{le2019pattern}. Shear, affine transformation, perspective transformation, contrast, brightness, and saturation augment are used in our tasks. The loss obtained by the data augmentation method is two orders of magnitude higher than that without using it, but the accuracy of single model on the validation set is approximate. However, the model voting based on models obtained from different data augmentation achieves large performance gains compared to single model.\\\\
\textbf{Multiple Resolutions.} The provided images in the training set are resized to $400\times 400$ by the organizer, so our baseline model is to use the original $400\times 400$ image. In addition, we have tried different resolutions, e.g., $300 \times 200, 320 \times 240, 320 \times 320, 360 \times 200, 360 \times 240, 400 \times 200, 400 \times 240, 400 \times 300, 400 \times 400, 480 \times 240, 500 \times 400, 600 \times 300, 600 \times 400, 800 \times 400$. \\\\
\textbf{Synchronized Batch Normalization (SyncBN).} Compared to traditional OCR tasks, table recognition often requires longer sequence decoding lengths and larger input image sizes, which will cause large GPU memory usage, and thus lead to insufficient batch size. Synchronized BatchNorm (SyncBN) is an effective type of batch normalization used for multi-GPU training. Standard batch normalization only normalizes the data within each device (GPU). SyncBN normalizes the input within the whole mini-batch.\\\\
\textbf{Feature Concatenation of Layers in Transformer Decoder (FeaC).}  Different with the original MASTER, we concatenate the outputs of the last two transformer layers and apply a linear projection on the concatenated feature to obtain the last-layer feature. The last-layer feature is used for the final classification.\\\\
\textbf{Model Ensemble.} Model ensemble is a very widely used technique to improve the performance on many deep learning tasks. Voting and bagging are two main ensemble strategy. Voting: Models trained with different resolutions and data augmentation methods are used for voting. Bagging: Sampling of the self-segmented data set is performed with put-back, sub-models are trained with the sampled samples and finally the sub-models are fused.


\subsection{Task2: Table Content Reconstruction}
In the TCR task, we also use Ranger optimizer, SyncBN, data augmentation, feature concatenation and model ensemble. Different from the TSR task, we need to reconstruct the structure and the content of a table image simultaneously into the form of LaTeX code. We try some different strategies in this sub-task.

\textbf{Multiple Resolutions.} We have used the following input sizes in the TCR task, e.g., $400\times 400, 440\times 400, 480\times 400, 512\times 400, 544\times 400, 600\times 400$. We will discuss the influence of resolutions to the final performance in the experimental section.\\\\
\textbf{Pre-train Model.} We find that TABLE2LATEX-450K~\cite{deng2019challenges} is a large-scale data set sharing the same target (table recognition to Latex) with this competition. However, the Number of tokens categories in TABLE2LATEX-450K is much larger than that the data in this competition, and the labeling of the table structure is slightly different from this competition. Therefore, we carefully extract samples containing only the target categories with this competition from the TABLE2LATEX-450K. Finally, we obtain a 58k data set from TABLE2LATEX-450K as our  data for the pretrain model. 


\section{Experiments}
\label{sec:experiments}
\subsection{Metric}
This competition offers several metrics for evaluation,

\textbf{Exact Match Accuracy:} Fraction of predictions which match exactly with the ground truth.

\textbf{Exact Match Accuracy @95\% similarity:} Fraction of predictions with at least 95\% similarity between ground truth.

\textbf{Row Prediction Accuracy:} Fraction of predictions with a count of rows equal to the count of rows in the ground truth.

\textbf{Column Prediction Accuracy:} Fraction of predictions with a count of cell alignment (``c'', ``r'', ``l'') tokens equal to the count of cell alignment tokens in the ground truth.

\textbf{Alpha-Numeric Characters Prediction Accuracy:} Fraction of predictions which has the same alphanumeric characters as in the ground truth.

\textbf{LaTeX Token Accuracy:} Fraction of predictions which has the same LaTeX tokens as in the ground truth.

\textbf{LaTeX Symbol Accuracy:} Fraction of predictions which has the same LaTeX Symbols as in the ground truth.
\textbf{Non-LaTeX Symbol Prediction Accuracy:} Fraction of predictions which has the same Non-LaTeX Symbols as in the ground truth.

The exact match accuracy is used for the final ranking.

\subsection{Implementation Details}
In the TCR training, 4 Tesla V100 GPUs are used with the batch size 8 in each GPU while the input image size is 400 × 400, and batch size differs accordingly when using different resolutions. The maximum sequence length is 500. In default, Synchronized BN~\cite{zhang2018context} and Ranger optimizer~\cite{lessw2019ranger} are used in our experiments. The initial learning rate of optimizer is 0.001 with step learning rate decay. The maximum sequence length is 250 for the TSR task, the other hyper-parameter settings of the TSR training are the same as the TCR training.
All models are trained based on our own FastOCR toolbox. FastOCR is a fast and powerful text detection, text recognition and key information extraction framework. Motivated by MMDetection~\cite{chen2019mmdetection}, our FastOCR also leverages the mechanisms of configuration and register. We recommend the readers to refer to one of our another report~\cite{ye2021ICDAR}.

\subsection{TCR}
The method and tricks of our table recognition system is described above. We will display some experiment results as follow.

\begin{table}[h]
\centering
\small
\begin{tabular}{|@{\ }c@{\ }|@{\ }c@{\ }|@{\ }c@{\ }|@{\ }c@{\ }|@{\ }c@{\ }|@{\ }c@{\ }|c|c|c|c|c|c|}
\hline
\textbf{Ranger} & \textbf{Pre-train} & \textbf{Augment} & \textbf{SyncBN} & \textbf{FeaC}  & \textbf{Ensemble} & \multicolumn{6}{c|}{\textbf{TCR metrics}} \\
\cline{7-12} & & & & &
& AA & LTA & LSA & SA & EM@95\% & EM 
\\ \hline

 &  &  &  &  &  & 0.8116 & 0.7026 & 0.9499 & 0.5446 & 0.6823 & 0.4710 \\ \hline

\checkmark &  &  &  &  &  &  0.8064 & 0.6958 & 0.9504 & 0.5378 & 0.6807 & 0.4731 \\ \hline

\checkmark & \checkmark  &  &  &  &  &  0.8200  & 0.7157 & 0.9535  & 0.5670 & 0.7078 & 0.4960 \\ \hline
\checkmark & \checkmark & \checkmark &  &  &  &  0.8283 & 0.7214 & 0.9540 & 0.5696 & 0.7047 & 0.5018 \\ \hline
\checkmark & \checkmark & \checkmark & \checkmark &  &   & 0.8356 &  0.7183 & 0.9572 & 0.5602  & 0.6995 & 0.4903 \\ \hline
\checkmark & \checkmark   & \checkmark & \checkmark  & \checkmark  &    & 0.8158 & 0.7099 & 0.9561 & 0.5670 & 0.7031 & 0.4966 \\ \hline
\checkmark & \checkmark   & \checkmark &  \checkmark   & \checkmark  & \checkmark   &  \textbf{0.8690}  &  \textbf{0.7511}  & \textbf{0.9577}  &  \textbf{0.6207} & \textbf{0.7386} & \textbf{0.5586} \\ \hline
\end{tabular}
\caption{End-to-end evaluation on the TCR test data set under six indicators. AA: Alpha-Numeric Characters Prediction Accuracy, LTA: LaTeX Token Accuracy, LSA: LaTeX Symbol Accuracy, SA: Non-LaTeX Symbol Prediction Accuracy, EM: Exact Match Accuracy, EM@95\%: Exact Match Accuracy @95\% similarity.}
\label{tab:endtoendEval}
\end{table}

\begin{table}[h]
\centering
\begin{tabular}{|c|c|c|c|c|c|c|}
\hline
\textbf{Resolution} & \multicolumn{6}{c|}{\textbf{TCR metrics}} \\ \cline{2-7}
& AA   & LTA  & LSA  & SA  & EM@95\%  & EM  \\ \hline
400 x 400   & 0.8200  & 0.7157 & 0.9535  & 0.5670 & 0.7078 & 0.4960  \\ \hline
440 x 400   & 0.8356  & 0.7209 & 0.9514 & 0.5618 & 0.6979 & 0.4997  \\ \hline
480 x 400   & 0.8335  & 0.7172 & 0.9514 & 0.5712 & 0.7047 & 0.5054 \\ \hline
512 x 400   & 0.8367  & 0.7282 & 0.9530 & 0.5665 & 0.7157 & 0.5080 \\ \hline
544 x 400   & 0.8231  & 0.7271 & 0.9535 & 0.5701 & \textbf{0.7198} & 0.5096 \\ \hline
600 x 400   & \textbf{0.8377}  & \textbf{0.7297} & \textbf{0.9561} & \textbf{0.5837} & 0.7146 & \textbf{0.5153} \\ \hline

\end{tabular}
\caption{Comparison of different resolutions on the TCR test data set under six indicators. AA: Alpha-Numeric Characters Prediction Accuracy, LTA: LaTeX Token Accuracy, LSA: LaTeX Symbol Accuracy, SA: Non-LaTeX Symbol Prediction Accuracy, EM: Exact Match Accuracy, EM@95\%: Exact Match Accuracy @95\% similarity.}
\label{tab:endtoendEvalResolution}
\end{table}

In this competition, we have conducted some evaluations and recorded the results. Our model is tuned based on the validation set. In Table~\ref{tab:tsreval}, we have shown the results of different settings on the final test data set. The results are collected from the official website.

According to the results, we have the following observations,
\begin{itemize}[leftmargin=.1in]
  \item   pre-trained model can significantly improve the model performance. We speculate that this is due to the relatively limited data set provided in this competition. 
  We want to point out that the model trained on TABLE2LATEX-450K has 0\% accuracy when evaluated on the validation set of this competition. This means that the labeling formats of these two data sets are slightly different. Regardless of the 0\% accuracy, model fine-tuning helps a lot.
  \item data augmentation improves the performance by around 0.5\%; feature concatenation boosts the performance by 0.6\%.
  \item model ensemble largely improves the single model. We observe models trained under different settings (e.g., resolution, data augmentation) can provide more complementary information.
  \item there is some inconsistency between the results on the validation set and on the test set. Ranger shows much better performance than Adam~\cite{kingma2014adam} on the validation set, but only slightly performs better on the test set. SyncBN shows better performance on the validation set, but seems to make the performance worse on the test set.
\end{itemize}

We also evaluate the influence of different resolutions to the performance. The results are shown in Table~\ref{tab:endtoendEvalResolution}. We can find that large resolution shows better performance. In the TCR task, large resolution is important for the algorithm to recognize each character.



\subsection{TSR}
Similar to TCR, we evaluate different settings. The results are shown in Table~\ref{tab:tsreval}.
\begin{table}[h]
\centering
\begin{tabular}{|c|c|c|c|c|c|c|c|c|c|}
\hline
\textbf{Ranger} & \textbf{SyncBN} & \textbf{Augment} & \textbf{FeaC}  & \textbf{Ensemble} & \multicolumn{4}{c|}{\textbf{TSR metrics}} \\
\cline{6-9} & & & &
& EM@95\% & RA & CA & EM
\\ \hline
\checkmark & \checkmark   &     &   &   & 0.8488 &  0.9369  &  0.8651 &  0.6922 \\ \hline
\checkmark & \checkmark   &  \checkmark  &   &  & 0.8365 & 0.9332  & 0.8656 & 0.6890 \\ \hline
\checkmark & \checkmark   &  \checkmark  & \checkmark  &  &   0.8483  &   0.9414  &  0.8674  & 0.6958 \\ \hline
\checkmark & \checkmark   &  \checkmark   & \checkmark  & \checkmark  &  \textbf{0.8765}  & \textbf{0.9496}  &  \textbf{0.8874}  &  \textbf{0.7444}  \\ \hline
\end{tabular}
\caption{End-to-end evaluation on the TSR test data set under four indicators. EM: Exact Match Accuracy, EM@95\%: Exact Match Accuracy @95\% similarity, RA: Row Prediction Accuracy, CA: Column Prediction Accuracy.}
\label{tab:tsreval}
\end{table}

We observe that from Table~\ref{tab:tsreval},
\begin{itemize}[leftmargin=.1in]
  \item feature concatenation improves the performance by around 0.7\%. 
  \item model ensemble is very effective also for the TSR task.
\end{itemize}

\begin{table}[h]
\centering
\begin{tabular}{|c|c|c|c|c|}
\hline
\textbf{Resolution} & \multicolumn{4}{c|}{\textbf{TSR metrics}} \\ \cline{2-5}
          & EM@95\%      & RA & CA   & EM   \\ \hline
300 x 200   &   \textbf{0.8615}  &   0.9359  &  \textbf{0.8806} &   0.7058  \\ \hline
320 x 320   &   0.8502  &   0.9337  &  0.8715  &   \textbf{0.7063}  \\ \hline
360 x 240   &   0.8556  &   0.9364  &  0.8710  &   0.7044  \\ \hline
400 x 200   &   0.8502  &   0.9314  &  0.8697  &   0.6985  \\ \hline
400 x 300   &   0.8597  &   0.9400  &  0.8724  &   0.7035  \\ \hline
400 x 400   &   0.8483  &   \textbf{0.9414}  &  0.8674  &   0.6958  \\ \hline
480 x 240   &   0.8511  &   0.9355  &  0.8669  &   0.6922  \\ \hline
500 x 400   &   0.8352  &   0.9291  &  0.8669  &   0.6804  \\ \hline
600 x 400   &   0.8361  &   0.9278  &  0.8633  &   0.6849  \\ \hline
800 x 400   &   0.8025  &   0.9187  &  0.8647  &   0.6463  \\ \hline

\end{tabular}
\caption{Comparison of different resolutions on the TSR test data set under four indicators. EM: Exact Match Accuracy, EM@95\%: Exact Match Accuracy @95\% similarity, RA: Row Prediction Accuracy, CA: Column Prediction Accuracy.}
\label{tab:endtoendEval}
\end{table}

In Table~\ref{tab:endtoendEval}, we report the results of TSR under different resolutions. We can see that, different from the TCR task, small resolution performs better than large resolution for the TSR task. This is due to that a large resolution allows the model to distinguish clearly the text content, while a small resolution allows the model to focus on the structure of the table.



\begin{figure}[t] 
    \centering
    \includegraphics[width=0.75\textwidth]{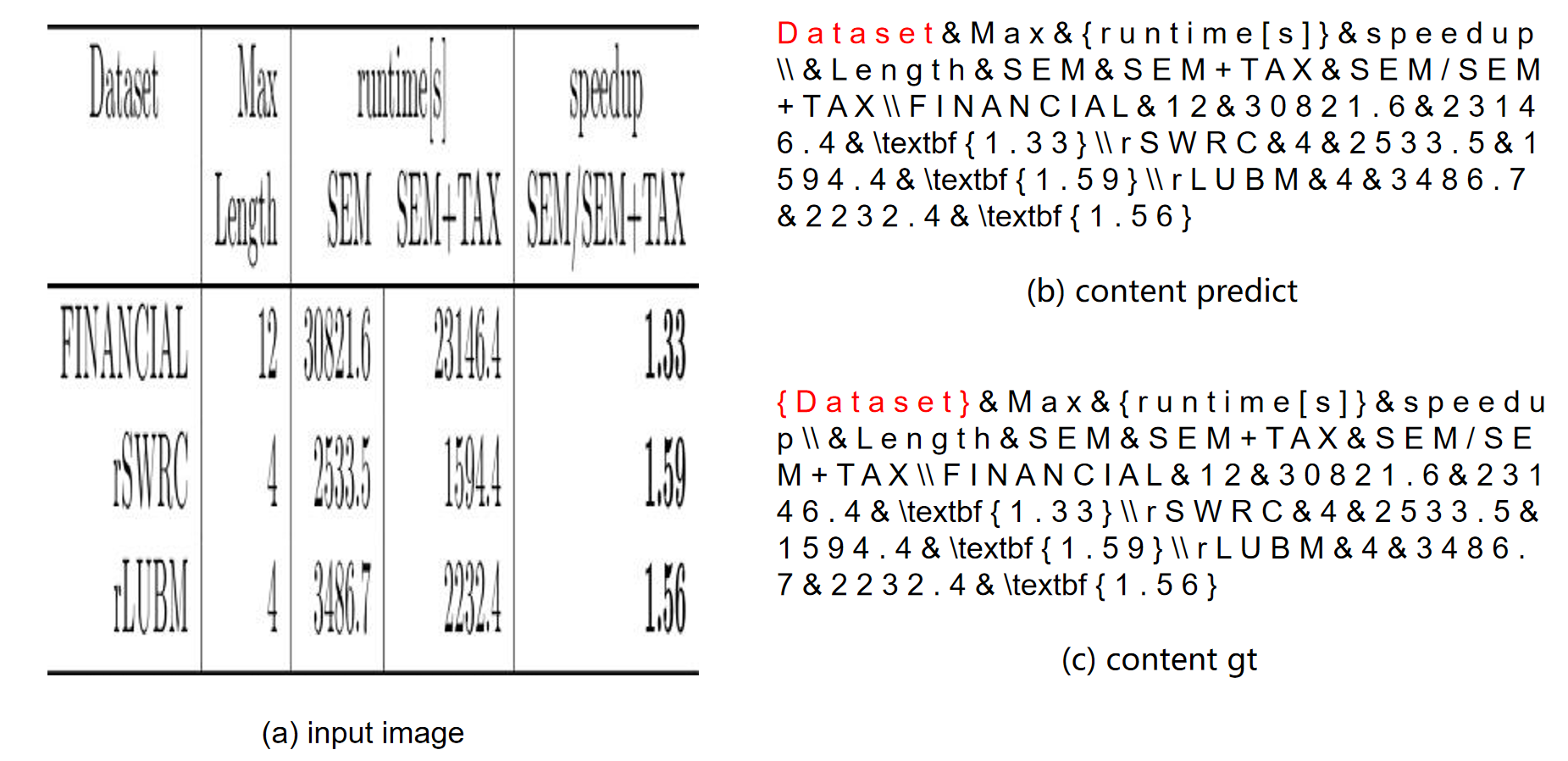}
    \caption{An example of wrong table content prediction.}
    \label{fig:wrongTCR}
\end{figure}

\begin{figure}[t] 
    \centering
    \includegraphics[width=0.75\textwidth]{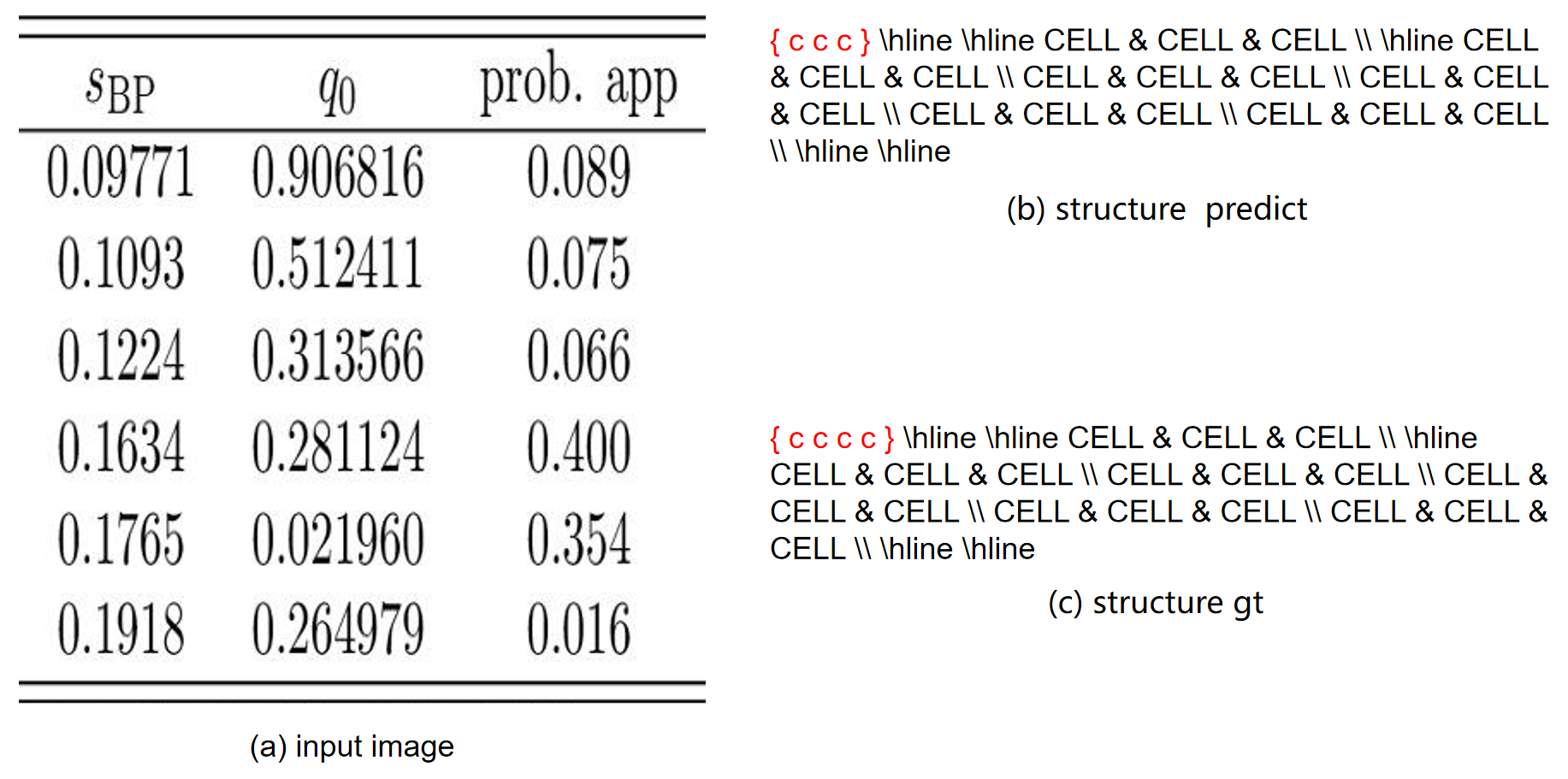}
    \caption{An example of wrong table structure prediction.}
    \label{fig:wrongTSR}
\end{figure}

\textbf{Discussion.} We have made some reflections from this competition,
\begin{itemize}[leftmargin=.1in]
  \item the data set of this competition is relatively small compared to some similar tasks~\cite{zhong2019image,Antonio2021ICDAR}. We believe that with increase of the data scale, the performance of the algorithm will improve dramatically.
  \item as shown in Figure~\ref{fig:wrongTCR}, the error is caused by the inconsistent data annotations. For instance, some classical inconsistency includes  with or without ``\{'' and ``\}'', with or without ``\textbackslash small'', confusion between ``\textbackslash textbf'' and ``\textbackslash mathbf'', confusion between ``\textbackslash em'' and ``\textbackslash emph'', and so on. 
  \item  as shown in Figure~\ref{fig:wrongTSR}, the most difficult part of the TSR task is to predict the head part (the structure part) of the table. 
  \item  the official images are resized to $400 \times 400$, which may cause heavy distortion, and thus seriously affect the performance of table content recognition. We believe that the padding-based pre-processing method is more suitable for these two sub-tasks.
\end{itemize}

\section{Conclusion}
\label{sec:conclusion}
In this paper, we present our solution for ICDAR 2021 Competition Scientific Table Image Recognition To LaTeX. Our model is based on MASTER. We optimize the MASTER model from several perspectives: network structure, optimizer, normalization method, pre-trained model, resolution of input image, data augmentation, and model ensemble. 
Our method achieves 0.7444 Exact Match and 0.8765 Exact Match @95\% on the TSR task, and obtains 0.5586 Exact Match and 0.7386 Exact Match 95\% on the TCR task.


\bibliographystyle{unsrt}  
\bibliography{references}

\begin{thebibliography}{10}

\bibitem{lu2019master}
Ning Lu, Wenwen Yu, Xianbiao Qi, Yihao Chen, Ping Gong, Rong Xiao, and Xiang
  Bai.
\newblock Master: Multi-aspect non-local network for scene text recognition.
\newblock {\em Pattern Recognition}, 2021.

\bibitem{Pratik2021ICDAR}
Pratik Kayal, Mrinal Anand, Harsh Desai, and Mayank Singh.
\newblock Icdar 2021 competition on scientific table image recognition to
  latex.
\newblock In {\em 2021 International Conference on Document Analysis and
  Recognition (ICDAR)}. IEEE, 2021.

\bibitem{lessw2019ranger}
Less Wright.
\newblock {Ranger-Deep-Learning-Optimizer}, 2019.

\bibitem{liu2019radam}
Liyuan Liu, Haoming Jiang, Pengcheng He, Weizhu Chen, Xiaodong Liu, Jianfeng
  Gao, and Jiawei Han.
\newblock On the variance of the adaptive learning rate and beyond.
\newblock In {\em Proceedings of the Eighth International Conference on
  Learning Representations (ICLR 2020)}, April 2020.

\bibitem{zhang2019lookahead}
Michael~R Zhang, James Lucas, Geoffrey Hinton, and Jimmy Ba.
\newblock Lookahead optimizer: k steps forward, 1 step back.
\newblock {\em arXiv preprint arXiv:1907.08610}, 2019.

\bibitem{yong2020gradient}
Hongwei Yong, Jianqiang Huang, Xiansheng Hua, and Lei Zhang.
\newblock Gradient centralization: A new optimization technique for deep neural
  networks.
\newblock In {\em European Conference on Computer Vision}, pages 635--652.
  Springer, 2020.

\bibitem{izmailov2018averaging}
Pavel Izmailov, Dmitrii Podoprikhin, Timur Garipov, Dmitry Vetrov, and
  Andrew~Gordon Wilson.
\newblock Averaging weights leads to wider optima and better generalization.
\newblock {\em arXiv preprint arXiv:1803.05407}, 2018.

\bibitem{le2019pattern}
Anh~Duc Le, Bipin Indurkhya, and Masaki Nakagawa.
\newblock Pattern generation strategies for improving recognition of
  handwritten mathematical expressions.
\newblock {\em Pattern Recognition Letters}, 128:255--262, 2019.

\bibitem{deng2019challenges}
Yuntian Deng, David Rosenberg, and Gideon Mann.
\newblock Challenges in end-to-end neural scientific table recognition.
\newblock In {\em 2019 International Conference on Document Analysis and
  Recognition (ICDAR)}, pages 894--901. IEEE, 2019.

\bibitem{zhang2018context}
Hang Zhang, Kristin Dana, Jianping Shi, Zhongyue Zhang, Xiaogang Wang, Ambrish
  Tyagi, and Amit Agrawal.
\newblock Context encoding for semantic segmentation.
\newblock In {\em Proceedings of the IEEE conference on Computer Vision and
  Pattern Recognition}, pages 7151--7160, 2018.

\bibitem{chen2019mmdetection}
Kai Chen, Jiaqi Wang, Jiangmiao Pang, Yuhang Cao, Yu~Xiong, Xiaoxiao Li,
  Shuyang Sun, Wansen Feng, Ziwei Liu, Jiarui Xu, et~al.
\newblock Mmdetection: Open mmlab detection toolbox and benchmark.
\newblock {\em arXiv preprint arXiv:1906.07155}, 2019.

\bibitem{ye2021ICDAR}
Jiaquan Ye, Xianbiao Qi, Yelin He, Yihao Chen, Dengyi Gu, Peng Gao, and Rong
  Xiao.
\newblock Pingan-vcgroup’s solution for icdar 2021 competition on scientific
  literature parsing task b: Table recognition to html.
\newblock {\em arXiv}, 2021.

\bibitem{kingma2014adam}
Diederik~P Kingma and Jimmy Ba.
\newblock Adam: A method for stochastic optimization.
\newblock {\em arXiv preprint arXiv:1412.6980}, 2014.

\bibitem{zhong2019image}
Xu~Zhong, Elaheh ShafieiBavani, and Antonio~Jimeno Yepes.
\newblock Image-based table recognition: data, model, and evaluation.
\newblock {\em arXiv preprint arXiv:1911.10683}, 2019.

\bibitem{Antonio2021ICDAR}
Antonio~Jimeno Yepes and Doug Burdick.
\newblock Icdar 2021 competition on scientific table image recognition.
\newblock {\em ICDAR}, 2021.

\end{thebibliography}

\end{document}